\documentclass[letterpaper, 10 pt,  journal, twoside]{IEEEtran/IEEEtran}

\IEEEoverridecommandlockouts

\usepackage{graphics} 
\usepackage{epsfig} 
\usepackage{amsmath} 
\usepackage{amssymb}  
\usepackage{algorithm}
\usepackage[noend]{algpseudocode}
\usepackage[noadjust]{cite}
\usepackage{balance}

\begin{document}

\markboth{IEEE Robotics and Automation Letters. Preprint Version. Accepted December, 2016}
{Abraham \MakeLowercase{\textit{et al.}}: Ergodic Exploration using Binary Sensing for Non-Parametric Shape Estimation}

\author{Ian Abraham$^{1}$, Ahalya Prabhakar$^{1}$, Mitra J.Z. Hartmann$^{1,2}$, and Todd D. Murphey$^{1}$
\thanks{Manuscript received: September, 10, 2016; Revised December, 10, 2016; Accepted December, 26, 2016. }
\thanks{This paper was recommended for publication by Editor Francois Chaumette upon evaluation of the Associate Editor and Reviewers' Comments. This work was supported by the National Science Foundation award IIS-1426961 to TDM and National Institute of Health award R01-NS093585 to MJZH and TDM.}
\thanks{Authors are with the Neuroscience and Robotics Laboratory (NxR) at the Department of Mechanical Engineering$^{1}$ and Biomedical Engineering$^{2}$, Northwestern University, 2145 Sheridan Road Evanston, IL 60208 USA.}
\thanks{{\tt\small Email: ianabraham2015@u.northwestern.edu, AhalyaPrabhakar2013@u.northwestern.edu, m-hartmann@northwestern.edu, t-murphey@northwestern.edu}}
\thanks{Digital Object Identifier (DOI): see top of this page.}}

\title{Ergodic Exploration using Binary Sensing for Non-Parametric Shape Estimation}
\bstctlcite{IEEEexample:BSTcontrol}
\maketitle

\begin{abstract}

Current methods to estimate object shape\textemdash using either vision or touch\textemdash generally depend on high-resolution sensing. Here, we exploit ergodic exploration to demonstrate successful shape estimation when using a low-resolution binary contact sensor. The measurement model is posed as a collision-based tactile measurement, and classification methods are used to discriminate between shape boundary regions in the search space. Posterior likelihood estimates of the measurement model help the system actively seek out regions where the binary sensor is most likely to return informative measurements. Results show successful shape estimation of various objects as well as the ability to identify multiple objects in an environment. Interestingly, it is shown that ergodic exploration utilizes non-contact motion to gather significant information about shape. The algorithm is extended in three dimensions in simulation and we present two dimensional experimental results using the Rethink Baxter robot.

\end{abstract}


\section{Introduction}

\IEEEPARstart{T}{actile} sensing is often associated with shape estimation \cite{moll2004reconstructing, meier2011probabilistic,lederman1987hand,huynh2010finding} and mapping problems \cite{tactileMap} in conditions where visual sensing may be limited. In some instances, tactile sensing is used to supplement vision-based sensing to improve shape estimates \cite{ilonen2014three,tactileTumor}. The richness of touch as a sensing modality is underscored by the development of novel tactile sensors \cite{ts_ref_1,ts_ref_2,ts_ref_3,ts_ref_4} for use in a myriad of applications ranging from robot-assisted tumor detection  \cite{tactileTumor}, to texture recognition, and feature localization \cite{hughes2014soft,hughes2015texture}. These advances in tactile sensor technology require corresponding advances in active exploration algorithms and the interpretation of tactile-based sensor data.  

Recent approaches for active exploration use random sampling-based search algorithms for sensor motion planning that require a separate controller for path following \cite{tactileAppear}. Other methods use task-specific probabilistic spatial methods for shape estimation that are updated based on sensor poses that minimize measurement uncertainty \cite{meier2011probabilistic, 7378871, ye2014real, dune2008active, lepora2013active}. A common approach in the literature is motion planning for sensing followed by feedback regulation of the generated plan \cite{matsubara2016active,mur2015probabilistic,lepora2013active, dune2008active}. Moreover, most methods focus on one object at a time. In contrast, the presented work integrates planning and control into a feedback law. As a result, the method uses sensor motion to actively sense for time-varying spatial information. We take the framework in \cite{miller2016ergodic} and use the feedback law developed here to enable real-time execution during shape estimation. Lastly, the feedback in the active sensing algorithm compensates for low resolution sensors and the specification of the algorithm is independent of the number of objects.

We show that ergodic exploration with Sequential Action Control (eSAC) can be used for active exploration with respect to time-varying tactile-based distributed information \cite{mathew2011metrics, miller2013trajectory}. Previous applications of ergodic theory utilize parametric measurement models for localization tasks \cite{miller2015optimalrange, miller2016ergodic}. The current work demonstrates localization and estimation of non-parametric shapes using a binary sensor model with classification methods. In contrast to other methods of active sensing for shape estimation \cite{matsubara2016active, Cootes1999567, rousson2005efficient, ye2014real, lepora2013active}, the proposed algorithm automatically encodes dynamical constraints without any overhead spatial discretization or motion planning. In addition, the algorithm incorporates sensor measurement information to actively adjust shape estimates and synthesize tactile-information based control actions. As a result, the algorithm automatically adjusts the control synthesis for multiple objects in an environment. Notably, ergodic exploration uses non-contact motion data (sensor motion not in contact with an object) \cite{wong2014not, koval2013pose} to improve the shape estimate. The idea of utilizing free space is often found in other related works of pose estimation and tracking \cite{wong2014not, koval2013pose, ganapathi2012real, newcombe2015dynamicfusion} and is emphasized in our work. As a final contribution, we show the algorithm is modular with respect to the choice of shape representation and tactile information distribution.

The paper outline is as follows: Section \ref{tactileInformation} motivates the use of a binary contact sensor and outlines the implications for tactile sensing. Section \ref{eSAC} describes the robot control algorithm for active exploration and motivates ergodicity as a measure for exploration. Section \ref{shapeEstimation} explains how shape estimation is achieved and how measurements update the shape estimate and the control policy. Experimental and simulated results and conclusion are shown in sections \ref{results} and \ref{conclusion} respectively.

\section{Tactile Information}\label{tactileInformation}

In biological systems, tactile sensing is generally an active process that incorporates feedback from multiple environmental cues \cite{gibson1962observations, chiel2009brain}. Humans, for instance, use fingers to grasp objects for manipulation and feature detection. Many rodents use hair-like appendages (whiskers) for tactile sensing \cite{guic1989rats,carvell1990biometric,hobbs2015spatiotemporal}. These biological sensors have a large set of actuators and sensor channels, thus the process control for active sensing is complex. We investigate a lower resolution version of tactile sensing that is simple to control for active sensing. Specifically, we show that a binary form of tactile sensing (i.e., collision detection) \cite{ts_ref_3, ts_ref_1}, has enough information for shape estimation, when combined with an active exploration algorithm that automatically takes into account regions of shape information. 

A binary tactile measurement model consists of a transition state from ``no collision'' to ``collision'' and vice versa.
We denote this measurement model as
\begin{equation}\label{eq:meas}
\Upsilon(x) =
  \begin{cases}
      \hfill 1,    \hfill & \phi(x) \leq 0 \\
      \hfill 0,    \hfill & \phi(x) > 0 \\
  \end{cases}
\end{equation}
where $x$ is the sensor state and $\phi(x)$ is a boundary function that determines a transition state if $\phi(x) \leq 0$ (output of 1). The goal of shape estimation is to determine $\phi(x)$.

\section{Active Exploration Using Ergodic Control}\label{eSAC}

\subsection{Motivation}

Typical algorithms used in active exploration and information maximization cast the problem of information acquisition as ``exploratory'' (wide spread search for diffuse information such as localization) or ``exploitative'' (direct search for highly structured information such as shape contours) \cite{low2008adaptive, krause2007nonmyopic}. Ergodic exploration \cite{mathew2011metrics, miller2012optimal, miller2013trajectory, miller2015optimalrange, miller2016ergodic} is responsive to both diffuse information densities and highly focused information. Thus, ergodic control seamlessly encodes wide spread coverage for diffuse information and localized search for focused information for both needs.

\subsection{Ergodic Metric}

A trajectory $x(t)$ of a sensor is ergodic with respect to a probability (``target'') distribution $\Phi(x)$ when the fraction of time the trajectory spends in an area within the search space is equal to the spatial statistics across the search space \cite{mathew2011metrics,shell2005ergodic}. The ergodic metric that measures this characteristic is given as \cite{mathew2011metrics,miller2013trajectory},
\begin{equation} \label{eq:erg_metric}
\mathcal{E} (x(t)) = \sum_{k \in {\mathbb Z}^n} \Lambda_{k} [c_{k}(x(t)) - \phi_{k}]^2
\end{equation}
where
\begin{eqnarray}
c_{k} (x(t)) & = & \frac{1}{t_f - t_0}\int_{t_0}^{t_f} F_k (x(\tau)) d\tau , \\
\phi_{k} & = & \int_{\mathcal{X}} \Phi({\bf x}) F_k ({\bf x}) d{\bf x} .
\end{eqnarray}
Here, $F_k (x)$ is a Fourier basis function, and $\Lambda_{k}$ are weights described in \cite{mathew2011metrics}. The target distribution $\Phi(x)$ is what drives the control synthesis for active-exploration.

Comparison between robot trajectory and distribution is done with equation (\ref{eq:erg_metric}) which takes the Fourier power series decomposition using (3) and (4) and directly compares the statistics of a trajectory with that of the spatial statistics $\Phi (x)$. As the ergodic metric is convex \cite{mathew2011metrics}, convex objectives subject to linear affine constraints are still convex. As a consequence, for linear affine dynamical systems, there are no local minimizers and must eventually search the whole space. A path is ``ergodic'' with respect to a distribution if the fraction of time the trajectory spends in a region in space is equivalent to the measure associated with that region. 

\subsection{Ergodic Sequential Action Control}

Ergodic control determines a discrete set of control inputs that aims to reduce the ergodic metric over time. Ergodic Sequential Action Control (eSAC) is a control scheme that generates ergodic control actions. Although trajectory optimization has been used to generate control synthesis for ergodic exploration \cite{miller2013trajectory,miller2016ergodic}, sequential action control provides a closed-form control that can immediately respond to changes in the ergodic metric in an infinite horizon setting \cite{ansari2015sequential}. In comparison with sample-based planners, SAC incorporates dynamical constraints. In particular, non-linear feedback control is desired when dealing with collisions and contact information that must be incorporated at the time of measurement acquisition. Thus, the motivating factor for using eSAC is a combination of compactness in its formulation and the ability to quickly react to measurements. 

Assuming control-affine dynamics of the form
\begin{equation}
\dot{x} = f(x(t), u) =  g(x) + h(x) u
\end{equation}
where $x \in \mathbb{R}^n$ and $u \in \mathbb{R}^m$, the trajectory cost is
\begin{equation}
J(x(t)) = q \mathcal{E} (x(t)) + \int_{t_0}^{t_f} \frac{1}{2} u(\tau)^T R u(\tau) d\tau
\end{equation}
where $q \in \mathbb{R}$ weights the ergodic metric and $R \in \mathbb{R}^{m \times m}$ is a positive definite matrix. During each sampling instance, eSAC computes the optimal action to reduce the cost function $J(x(t))$ in the following steps:
\subsubsection{Prediction}
eSAC forward simulates the system dynamics starting from the sampled sensor state $x_0$ at time $t_0$ to a finite time horizon $t_0 + T$ assuming a default control $u_0$ similar to NMPC methods \cite{grune2011nonlinear, chen1997quasi}. Cost sensitivity is computed by backwards simulating an adjoint variable, $\rho \in \mathbb{R}^n$ that satisfies the following differential equation
\begin{eqnarray}
\dot{\rho}  &=&  2 \frac{q}{T} \sum_{k \in \mathbb{Z}^n} \Lambda_k (c_k (x(t))  -  \phi_k) \frac{\partial}{\partial x} F_k (x(t)) \nonumber \\
& &\ \ \ \ \ \ \ \ \ \ \ \ \ \ -  \frac{\partial}{\partial x} f(x, u_0)^T \rho \\
&&\text{subject to } \rho(t_0 + T) = \vec{0} \nonumber .
\end{eqnarray}
The pair $x(t)$ and $\rho(t)$ are used to compute the optimal control action.

\begin{figure*}[!t]
\includegraphics[scale=1]{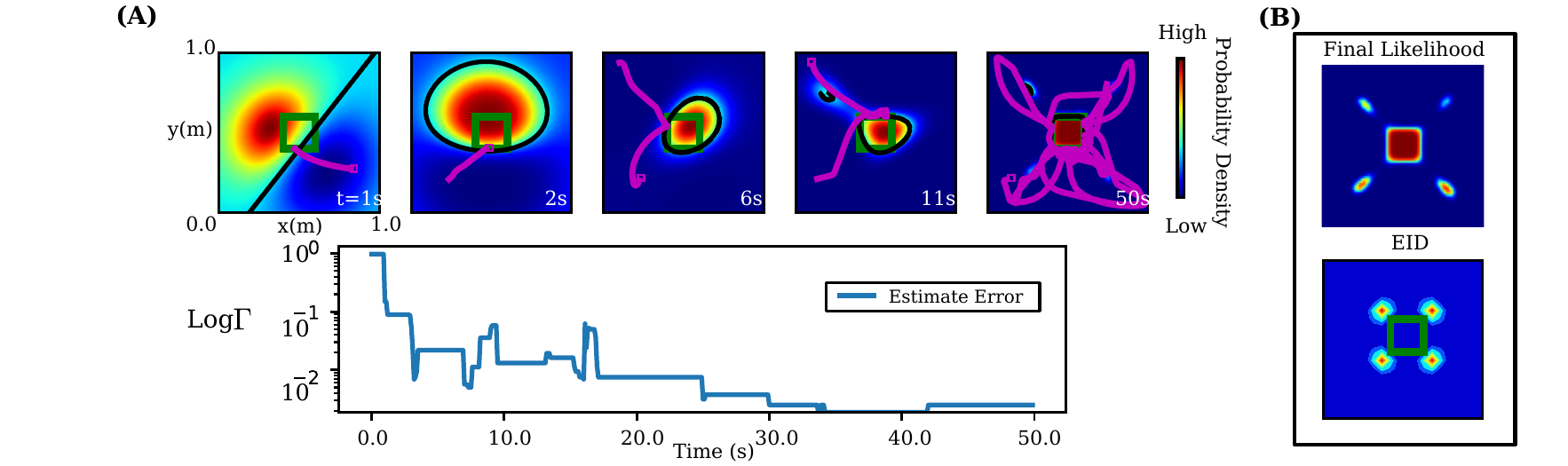}
\caption{ {\bf (A)} Simulated time sequence of tactile-based eSAC estimating a square (green) in $\mathcal{X} \subset \mathbb{R}^2 \subset \left[0,1 \right] \times \left[0,1 \right] $. Robot trajectory (magenta) translates in time from the previous final state (shown as a circle) to the current initial condition (shown as a magenta square). The boundary of the surface estimate is shown as a black line and the posterior likelihood is shown in the density plot. High likelihood of acquiring a collision is shown in red and low likelihood is shown in blue.  {\bf Bottom Row}: $\Gamma$ metric is defined as $\Gamma =  \int_x  \left(\phi_{\text{emp}}(x) - \phi_{\text{actual}} (x) \right) ^2 dx $ the integrated difference of the shape and the estimated shape squared. The $\Gamma$ measure drops quickly after first contact and then remains at an equilibrium as the shape estimate is updated. {\bf (B)} Note the posterior at the end of the simulation (top subplot) resembles the Expected Information Density (EID) (see \cite{miller2016ergodic} for derivation of EID) of the parametric model of the square shape (bottom subplot). }
\label{squareErgConv}
\end{figure*}

\subsubsection{Calculate Optimal Control Actions}

eSAC produces a schedule of control actions $u^* (t) \in [t_0, t_0 + T]$ that optimizes
\begin{eqnarray}
J_u = \frac{1}{2} \int_{t_0}^{t_0 + T} \left[ \frac{dJ_{\text track}}{d \lambda} - \alpha_d \right]^2 + \|u(t)\|_R^2 dt \\
\text{where } \frac{dJ_{\text track}}{d \lambda} = \rho(t)^T (f(x(t), u) - f(x(t), u_0)).
\end{eqnarray}
The term $\frac{dJ_{\text track}}{d \lambda}$ computes the rate of change of the cost with respect to a switch of infinitesimal duration $\lambda$ \cite{ansari2015sequential, tzora2016Model}. The value $\alpha_d \in \mathbb{R}^-$ dictates the aggressiveness of the control (more negative values tend to saturate the control). In this work a value for $\alpha_d = -555$ was used.
The closed form solution of $u^*$ is then given as
\begin{equation}
u^* = (\Lambda + R^T)^{-1} [\gamma u_0 + h(x)^T \rho \alpha_d]
\end{equation}
where $\gamma  \triangleq h(x)^T \rho \rho^T h(x)$. Additional information on the derivation of SAC can be found in \cite{ansari2015sequential}. In this work, we used values of $q=30$, $R=\text{diag} ( [ 0.01, 0.01 ] )$, $T=0.8$ for eSAC.

%
\begin{algorithm}
\caption{eSAC} \label{alg:eSAC}
\begin{algorithmic}[1]
\State \textbf{given} $x_0, c_{k,0}, \phi_{k,0}, t_{curr}, T, t_s, i=0$
\State $(x(t), \rho(t)) \gets $ prediction($x_0, t_{curr}, T$ )
\State $u^* (t) \gets $ calcControl($x(t), \rho(t)$)
\State \textbf{return} $u^*(t) \in \left[t_{curr}, t_{cuss}+t_s \right]$

\end{algorithmic}
\end{algorithm}

\section{Shape Estimation}\label{shapeEstimation}

\begin{figure}[h]
\includegraphics[scale=1]{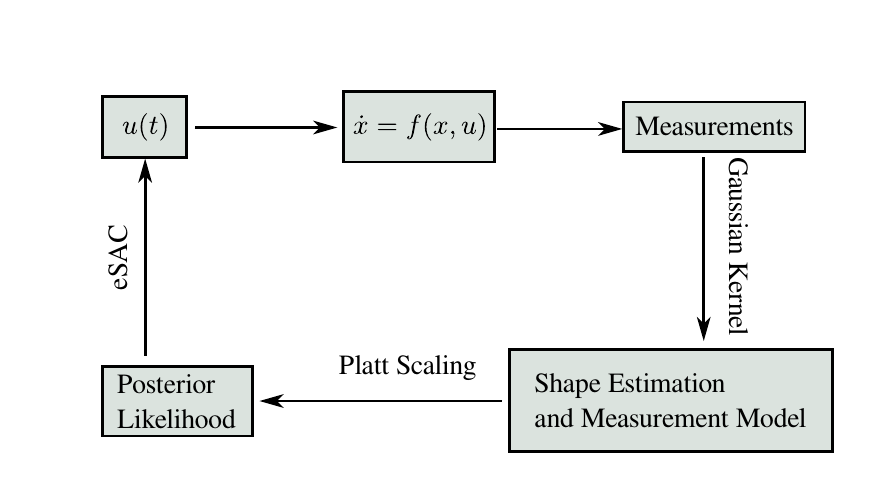}
\caption{ Block diagram for shape estimation. The measurements are processed by a Gaussian Kernel and made into a shape estimate. With Platt Scaling, the shape estimate is transformed into a posterior estimate of the likelihood of acquiring a specific binary sensor measurement, which then is converted into control using eSAC. }
\label{shapeReconFlow}
\end{figure}

The process for shape estimation is described in Fig. \ref{shapeReconFlow}. Shape estimates are obtained through repeated samples of the search space using directed motion of the sensor, as calculated by eSAC. The samples are processed using kernel basis functions in order to create decision boundaries. The decision boundaries are then processed into a posterior likelihood estimate using a common method in machine learning to generate statistics on a classification fit known as Platt Scaling \cite{vapnik1998statistical, chateau2007real, niculescu2005predicting, wu2004probability}. The returned spatial statistics on the fit is used for active sampling of the search space, thus updating the shape estimates. The process iterates at a user-specified sampling frequency. We further explain the process in the following sections.

\subsection{Binary Measurements for Shape Estimation}

Given a set of measurements $y_k \in \left[0, 1 \right]$ at time indexed by $k$ sampled at the corresponding set of sensor state $x_k$, shape estimation is accomplished by probabilistic classification methods \cite{lepora2013active, ye2014real, Cootes1999567, rousson2005efficient}.
Using the set of measurements, a decision boundary for the object is approximated as
\begin{eqnarray}
\phi(x) \approx \sum_{k} \alpha_k y_k K(x_k, x) + b
\end{eqnarray}
where $K(x_k, x)$ is the kernel basis function that determines the basis shape of the decision boundary and the parameters $\alpha_k$ and $b$ are optimized parameters based on the set of $y_k$ and $x_k$. The choice of kernel basis function determines the shape of the decision function \cite{scholkopf2001learning}. Arbitrary shape estimation is desired so a Gaussian kernel basis (also known as a radial basis function) is chosen \cite{cristianini2000introduction}. However, any kernel basis function can be used as a design choice. 
The Gaussian Kernel is given as
\begin{equation}
K(x_k,x) = e^{-\frac{(x_k - x)^2}{\sigma^2}} .
\end{equation}
A kernel of this form maps data into infinite dimensional feature space which provides flexibility for decision boundaries \cite{buhmann2000radial}. (In this work, the python package {\it Scikit-Learn} \cite{scikit-learn} is used for optimizing the boundary fit.)

\begin{figure*}[!ht]
\centering
\includegraphics[scale=0.9]{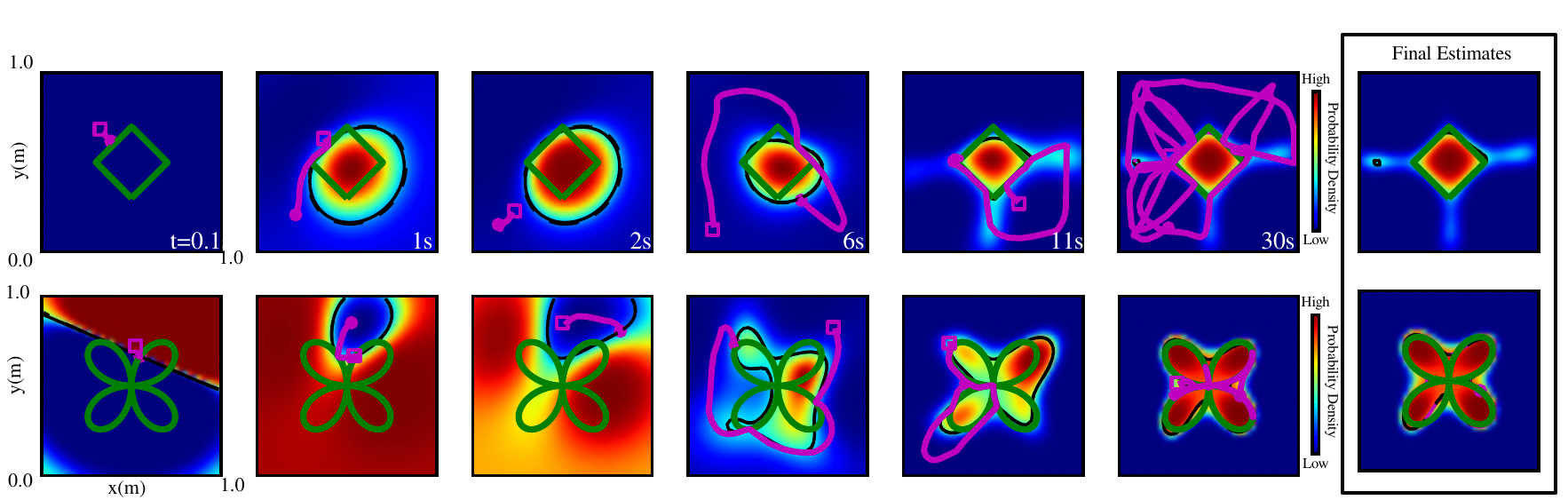}
\caption{Time series shape estimation of diamond and clover shape. The posterior likelihood estimate is shown at $0.1, 1, 2, 6, 11$, and $30$ seconds. Final shape estimates at $30s$ is shown in the enclosed box. Likelihood of measuring a collision measurement is denoted by the contours. The sensor trajectory (shown in magenta) traverses from the previous time window onto the current posterior. Shape estimates are depicted in the black line. Actual continuous contact motion does not begin until around $6$ seconds into the simulation. Nonetheless, through non-contact motion, the algorithm is able to estimate the shapes. After continuous contact occurs, the algorithm refines the shape estimate. The final shape estimates are highlighted in the enclosed rectangular box. Note that the shape estimate (denoted as the black line) for the square is underneath the actual shape. With the clover, part of the shape estimate is outside the shape boundary. }
\label{diffShapes}
\end{figure*}

\subsection{Posterior Likelihood Estimate (Platt Scaling)} \label{EID_Platt}

As the measurement model is unknown, we design the controller to be ergodic with respect to the likelihood distribution of acquiring a collision measurement. The distribution is updated for each measurement via Platt Scaling, defined as
\begin{equation}
P( y_k = 1 | x) = \frac{1}{1 + e^{A \phi(x) + B}} ,
\end{equation}
where $A$ and $B$ are solved through a regression fit and $\phi(x)$ is the current shape estimate.

\begin{figure}[h]
\centering
\includegraphics[scale=1.0]{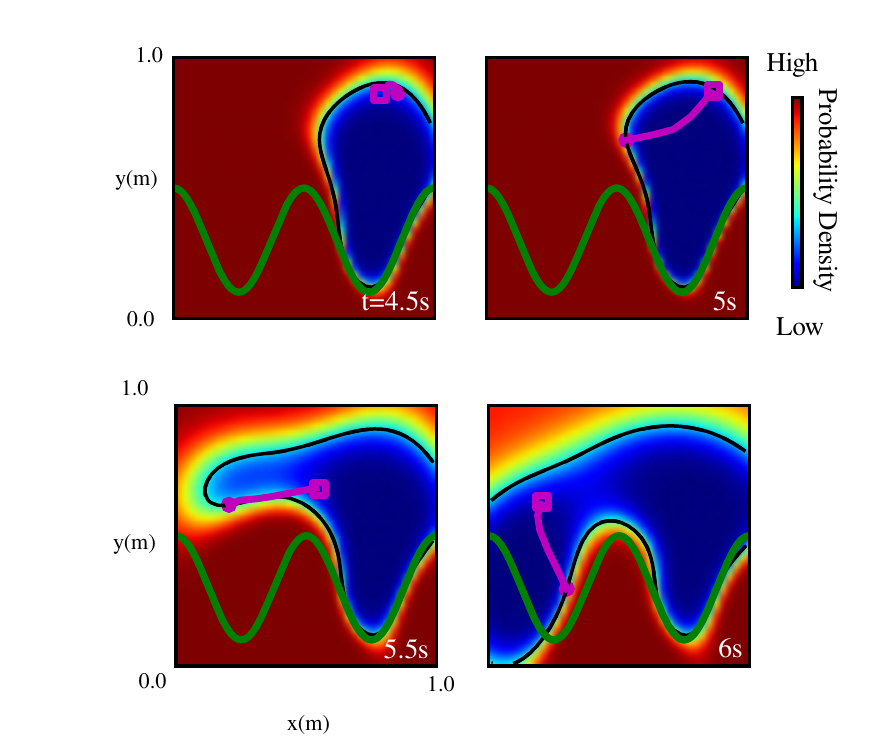}
\caption{Time series shape estimation of a spatial sine wave. High likelihood probabilities of a collision are shown in red and low probabilities in blue. Start point of sensor shown as the magenta square. By following the trajectory from $4.5s$ to $5.5s$, the shape estimate depicted by the black line is updated by the sensor even without a collision. At $6s$ the middle peak of the sine wave has been estimated and the sensor trajectory has not collided with the sine wave boundary.  }
\label{sineSurface}
\end{figure}

\begin{figure}[!th]
\centering
\includegraphics[scale=1]{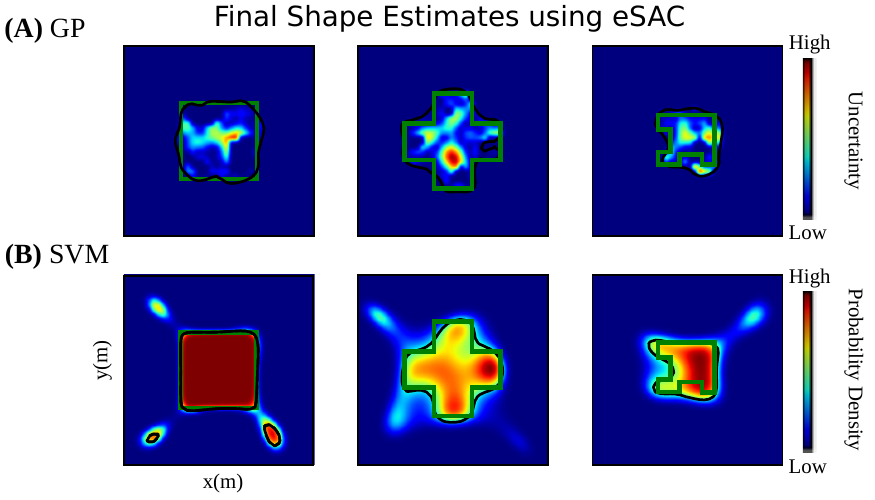}
\caption{Modularity in the eSAC algorithm is shown with the example shapes used in \cite{matsubara2016active} and their approach for generating shape estimate and a target distribution. (A) A Gaussian Process (GP) is used for shape representation and the uncertainty in the GP fit as the target distribution (this implies the robot will search near regions of high uncertainty in the fit). (B) Support Vector Machine (SVM) approach for shape representation using the posterior likelihood as the target distribution. Both methods using eSAC provide similar results in final shape estimates. }
\label{gp_svm}
\end{figure}

\section{Results}\label{results}

\subsection{Simulation Results}
Active shape estimation in $\mathbb{R}^2$ is done with double integrator dynamics given as
\begin{equation}
\dot{{\bf x}} = f({\bf x},u) =
\begin{bmatrix}
\dot{x}, &
\dot{y}, &
u_1, &
u_2
\end{bmatrix}^T .
\end{equation}
The binary measurements are collisions detected during the crossing of the boundary function $\phi(x) = 0$. Estimation in $\mathbb{R}^3$ is done with a similar double integrator model with dynamics
\begin{equation}
\dot{{\bf x}} = f({\bf x},u) =
\begin{bmatrix}
\dot{x}, &
\dot{y}, &
\dot{z}, &
u_1, &
u_2, &
u_3
\end{bmatrix}^T .
\end{equation}
No assumptions are needed on the location of the object and we initialize with a uniform distribution as the target distribution.

\begin{figure*}[!th]
\centering
\includegraphics[scale=1]{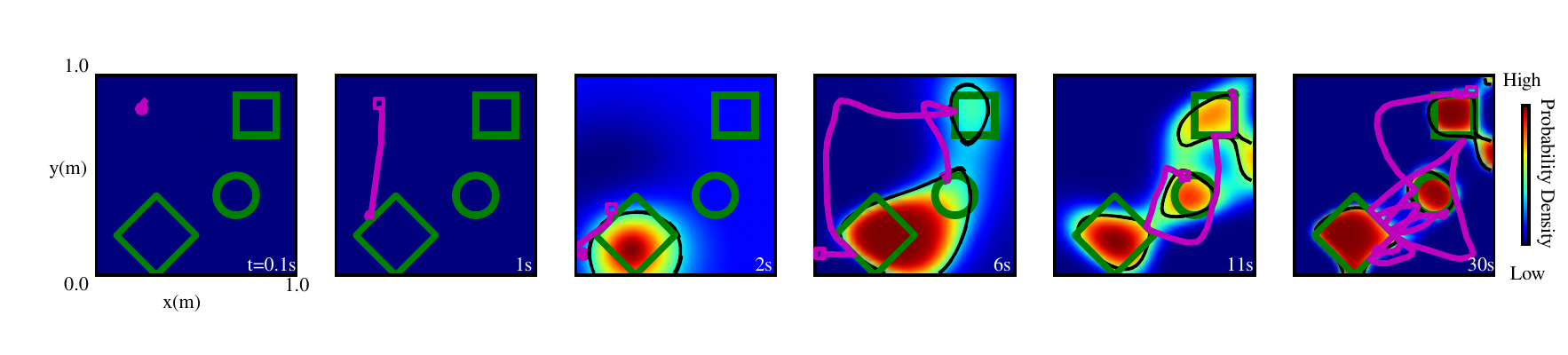}
\caption{Time series of eSAC exploring multiple objects. The posterior is viewed at times $0.1, 1, 2, 6, 11, 30$. Trajectories start at the magenta square. In each subplot, the sensor trajectory (magenta) starts from the endpoint of the previous time instant. Although the robot has no prior knowledge of the number of objects in the sensor state, the algorithm is still able to correctly estimate all the objects. Therefore, regardless of the number of shapes, the algorithm is still able to estimate all the shapes in the environment given a large enough time horizon.}
\label{mult_obj}
\end{figure*}

\begin{figure}[!h]
\centering
\includegraphics[scale=0.6]{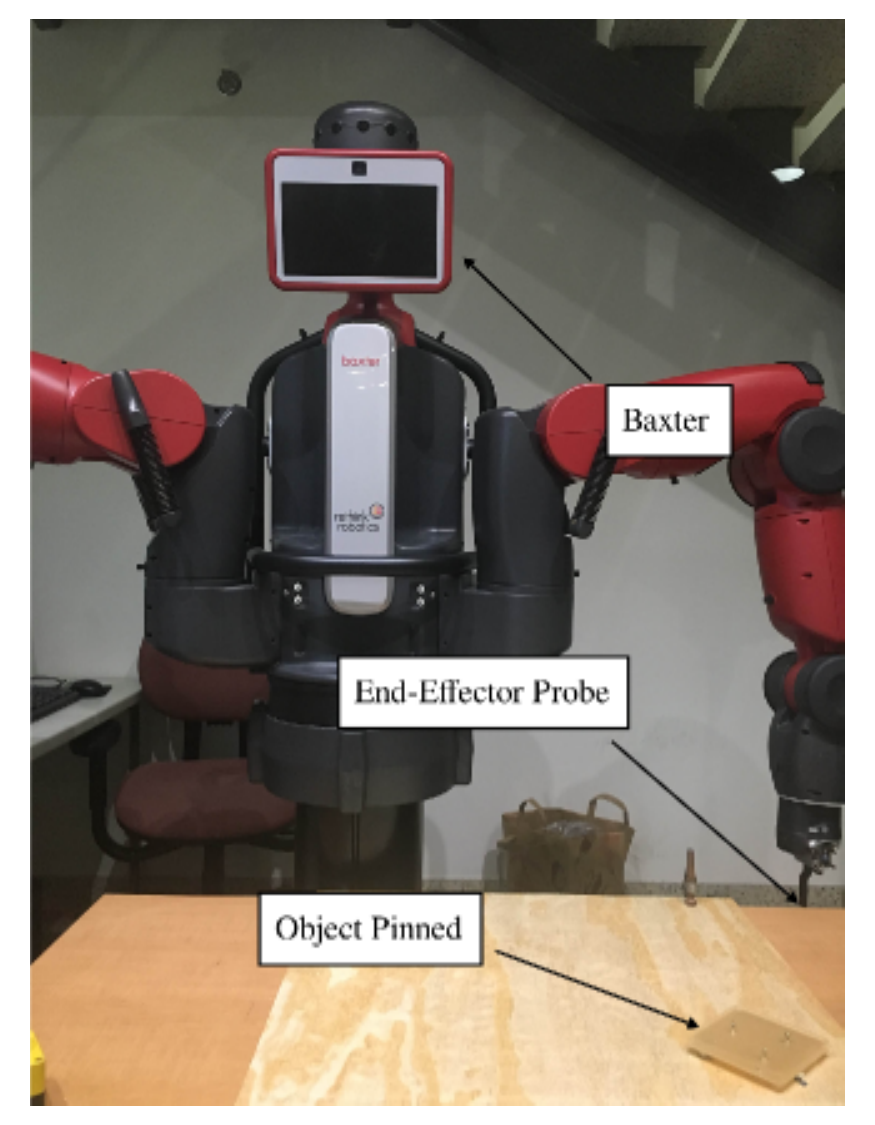}
\caption{ Baxter Robot used for experiments. An object is pinned down to the table for shape estimation using the end-effector probe. The probe itself has no sensors. Thus, the measured joint torques are used to detect collisions based on rapidly changing measurements at specific joint locations near the end-effector.  }
\label{baxter}
\end{figure}

First, we demonstrate that the proposed algorithm's shape estimate converges to the actual shape. Figure \ref{squareErgConv} shows time evolution of the posterior as well as the resulting shape estimate. The estimate depicted by the black line in Fig. \ref{squareErgConv} eventually converges to the shape, although much of the shape restructuring is accomplished from non-contact motion up until $6s$. Moreover, by comparing the posterior and the windowed sensor path, it is shown in Fig. \ref{squareErgConv} that the sensor path is drawn to high likelihood densities. This ensures that the posterior estimate is verified and updated as new sensor information is acquired. Note that the sensor is unable to access the interior of the shape, resulting in high likelihood probabilities within the shape. The posterior likelihood of the square shape estimate shows high likelihood probabilities near the corners of the square. As an aside, we take note of the similarities that the resultant likelihood has with the  expected information density (EID) of the known shape measurement model (see \cite{miller2016ergodic} for EID derivation). Specifically, high likelihood near the corners correspond to the similar large expectation of information for a square shape. If we define the measurement model of the known square as $\Upsilon(\bold{x}, x)$ where $x$ is search space and $\bold{x}$ is the robot's state space, then a measure of information is the Fisher Information Matrix \cite{cover2012elements} defined by
\begin{equation}
I (\bold{x}, x) = \frac{\partial \Upsilon}{\partial x} ^T \Sigma^{-1} \frac{\partial \Upsilon}{\partial x},
\end{equation} 
where $\Sigma$ is the measurement covariance. Assuming the measurement model is of the same form as in equation (\ref{eq:meas}) then the region with the largest information is at the corners where the slopes of the edges collide. Thus, large likelihood estimates should exist near corners if they have not been previously searched. 

\begin{figure}[h]
\centering
\includegraphics[scale=0.75]{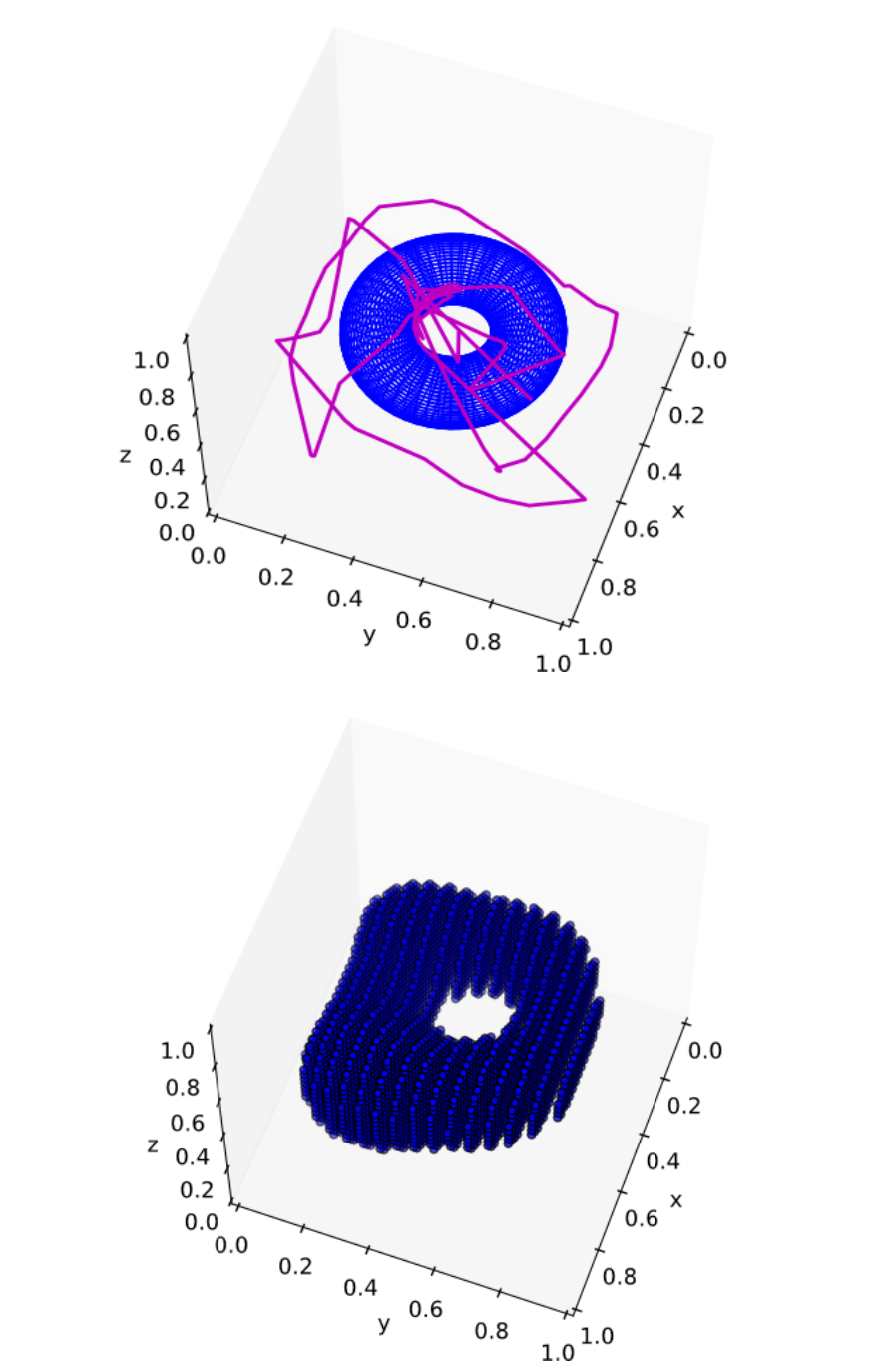}
\caption{eSAC trajectory for a Torus $\mathbb{S}^2$ in $\mathcal{X} \subset \mathbb{R}^3 \subset \left[0,1 \right] \times \left[0,1 \right] \times \left[0,1 \right]$ search space. {\bf Top}: In magenta, the robot trajectory on a Torus. {\bf Bottom}: Density plot of the shape estimate. Although the simulation was run for $40s$, the use of eSAC is shown to capture the doughnut shape inner circle feature which defines the torus shape. Utilizing an array of tactile sensors would reduce the error on the outer ring of the torus as larger coverage is needed for $ \mathbb{R}^3$. }
\label{torus}
\end{figure}

\begin{figure*}[!th]
\centering
\includegraphics[scale=1]{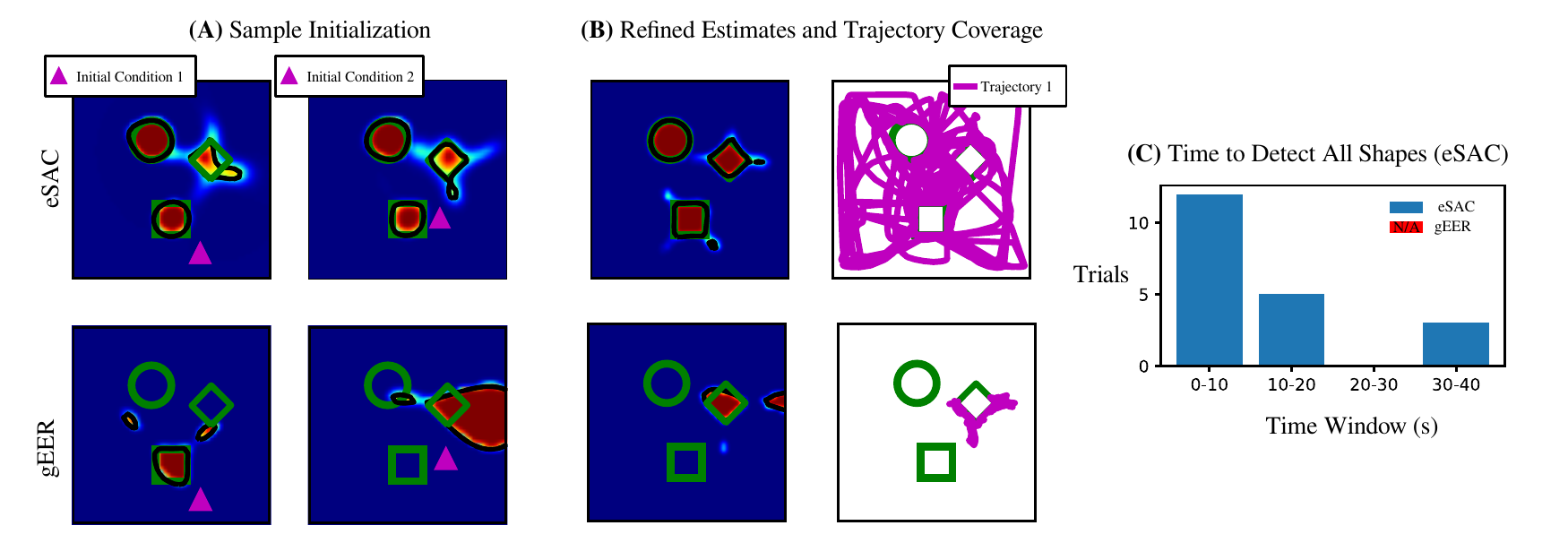}
\caption{Comparison of gEER and eSAC for detecting and refining multiple shape estimates. We randomly selected $20$ initial conditions with uniform distribution in the search space for both gEER and eSAC methods. (A) Trial samples shown from both eSAC and gEER. eSAC consistently detects all objects in the search space. (B) Refined estimates of the shape after detection. eSAC is shown to coarsely explore the whole region while spending a larger fraction of time around the shape estimates. gEER is driven by larger values which reduce uncertainty which is susceptible to local minima. (C) eSAC is shown to detect all shapes in the environment within $40s$ of simulation. In all $20$ trials, gEER does not successfully detect all shapes in the environment.} 
\label{e_greedy_esac}
\end{figure*}

\begin{figure}[!t]
\centering
\includegraphics[scale=1.0]{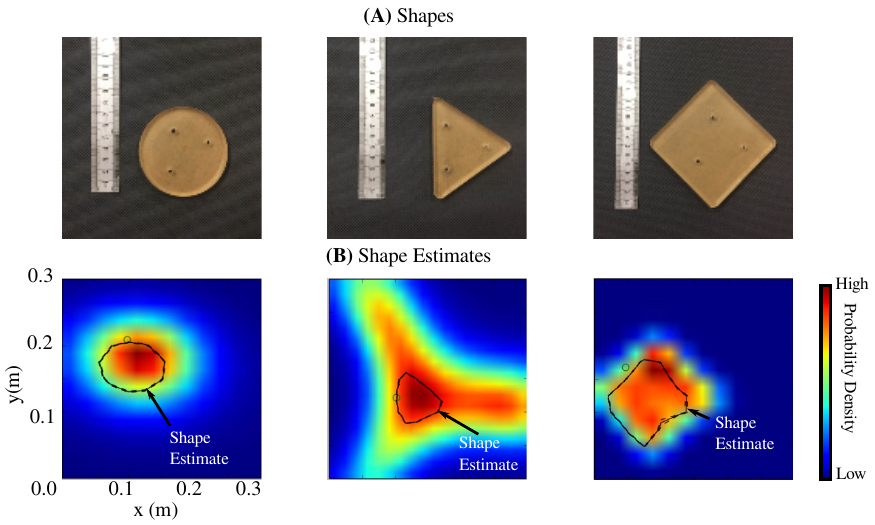}
\caption{Baxter experimental results for shape estimation. {\bf (A)} shows the shapes used for estimation. {\bf (B)} shows the location of Baxter's probe at the end of the estimation as well as the estimate of the shape as the dotted black line. The posterior at the end of the experiment is shown in the contour map. Note the prominent posterior likelihood in the triangle estimation around the edges are colored with high likelihood estimate, indicating that the estimated measurement model retains information about corners that is useful for localization. Error in shape area \cite{matsubara2016active} is within $2.5 \%, 7.5 \%,$ and $10.0 \%$ for the circle, triangle, and square respectively. (Coarse discretization of the distribution was created to generate the figure to prevent run-time degradation on Baxter which results in an off-centered visual error of $1cm$ in the distribution.) }
\label{baxterExp}
\end{figure}

The non-contact motion shown in Fig. \ref{sineSurface} is beneficial for tactile-based exploration. The sensor trajectory from time $4.5s$ to $6s$ updates the shape estimation of the sine wave boundary without measuring a collision for $t \in [4.5,6s]$. Knowing that no contact has occurred indicates that the shape is not in the current sensor state, but likely in other unexplored regions of sensor state. (Interestingly, tactile sensor arrays appear often in biological systems such as the rat whisker system \cite{hartmann2001active}). The algorithm is shown in Fig. \ref{diffShapes} to estimate both a clover shape and a diamond shape. Between times $2s$ to $6s$ in Fig. \ref{diffShapes}, non-contact motion is shown to drive the sensor trajectory towards an enclosing path that estimates the location of the center of the shape. After a few seconds, the sensor regularly is in contact with the shape as the estimate converges. The final estimates of the diamond and clover are shown by black lines in the rectangular box in Fig. \ref{diffShapes}. 

The final time posterior shown in Fig. \ref{squareErgConv} and Fig. \ref{diffShapes} show a resemblance in the posterior likelihood near the edges of the shape. This similarity in posterior estimates shows the algorithm approximates regions where there is high expected tactile information.

Shown in Fig. \ref{mult_obj}, multiple objects are allocated randomly in the sensor state. The sensor has no prior knowledge of where the shapes are, nor the number of shapes. The sensor contacts the first shape around $1s$ (Fig. \ref{mult_obj}), then between $2$ and $6s$ the sensor comes into contact with the remaining two shapes. Following the trajectory (magenta) traversed from time $11-30s$ shown in Fig. \ref{mult_obj}, the sensor distributes the time spent amongst the three shapes. Because the ergodic metric is convex with respect to the information distribution and thus convex with respect to the shape, unexplored regions of sensor state must be explored \cite{mathew2011metrics}. 

We further expand upon multiple shape detection with a comparison with a version of Greedy expected entropy reduction (gEER) exploration algorithm used in \cite{miller2016ergodic,kreucher2005sensor, fox1998active, feder1999adaptive, souza2014bayesian}. The algorithm samples nearby states centered at the robot probe's current location for regions with the largest probability of acquiring a collision measurement. The algorithm then moves the probe to that location, sampling along the way to adjust the shape estimate. In Fig.~\ref{e_greedy_esac}, we run $20$ trials of uniformly sampled initial conditions for both eSAC and gEER. Figure \ref{e_greedy_esac}(A) shows sample shape estimates for $80s$ simulations of both algorithms. eSAC is shown to detect all the shapes and begin shape refinement while gEER algorithm at most detects two shapes, but refines only one. We can see in (C) that eSAC detects all shapes within $40s$ of simulation time whereas the gEER detects two shapes at most. The time it takes to detect objects does depend on the distance of the shape and the control saturation of the robot probe which we maintained constant with both algorithms. In (B), we see the difference in the algorithm's area coverage. In particular, eSAC covers most of the search space coarsely with densely collected collision measurements around the shape. 

Modularity with respect to shape representation is demonstrated by comparing with the shapes used in \cite{matsubara2016active} with the Gaussian Process (GP) for shape estimation (see Fig.~\ref{gp_svm}). Here, eSAC is shown to work with both a GP and SVM for shape estimates. In particular, in Fig.~\ref{gp_svm}(A), eSAC uses the touch point selection metric used in \cite{matsubara2016active} for control synthesis which is based on the covariance of the GP fit. In comparison with eSAC, the work done in \cite{matsubara2016active} uses an Markov Decision Process (MDP) in order to control a robot manipulator towards acquiring touch data for shape estimation. In \cite{matsubara2016active}, a grid is defined and a path planner is used to traverse the grid. The large overhead state discretization and motion planning code that is necessary for the MDP formulation is not required for eSAC. Moreover, eSAC automatically encodes robot dynamic constraints. Notably, eSAC can be used with other visual-based estimators that feed initial shape estimates such that tactile data is used to refine the estimate.

This algorithm can be trivially extended to $\mathbb{R}^3$ as shown in simulation. Figure \ref{torus} shows the sensor trajectory in the top plot (magenta) actively sensing the torus shape. Shape estimation of the torus is shown to be successful in the lower sub-plot of Fig. \ref{torus} resulting in scalability to objects in $\mathbb{R}^3$. Performance for estimation in $\mathbb{R}^3$ depends on the robot dynamical constraints as well as the sensor area coverage.

\subsection{Experimental Results}

Experiments are done using the Baxter robot (Fig. \ref{baxter}). Binary measurements are taken by the end-effector probe during a large change of joint torques, indicating a collision has occurred. During collision, controller weight $R$ prevents the robot from dragging along the object. In the case that the control weight $R$ is not significant, heuristics are used to prevent overexertion. Baxter's end-effector is used as a probe for the $\mathbb{R}^2$ search space. No assumptions are needed on the location of the object or the height of the object.

Experimental results show that shape estimation can be accomplished with a robotic system with multiple sensors (Fig. \ref{baxterExp}). Although there is odometry error within the robot's joint states, shape estimations are still visually seen to match. Within the posterior regime seen in Fig. \ref{baxterExp}, the probability densities for the shapes with corners are shown to have unique features. In particular, the posterior for the triangle in Fig. \ref{baxterExp} extends outwards along the three corners indicating high probability of obtaining a collision measurement. Further measurements surrounding the edge regions reduce the likelihood of acquiring estimate which can be seen in with results from estimating the square shape.

\section{Conclusions and Future Work}\label{conclusion}

In this paper, an algorithm is presented for shape estimation using a binary sensor combined with ergodic exploration. It is shown that a binary measurement sensor can be treated as a lower resolution tactile sensor for shape estimation with active sensing. In addition, using ergodic exploration as the principle for active sensing uses sensor motion to capture shape features during estimation. The algorithm is shown to estimate shapes in $\mathbb{R}^2$ and $\mathbb{R}^3$ as well as an any number of shapes in the sensor domain. The algorithm is shown to be modular with respect to choice of shape representation. The algorithm is shown to have potential applications for active sensing with low resolution sensors with respect to shape estimation as well as applications in localization.

Future research directions include the use of multiple sensors for estimation. Another direction is to augment the current algorithm to include localization using the estimated measurement model. Thus, simultaneous localization and shape estimation should be possible.   
%

%

\bibliographystyle{IEEEtran}
\bibliography{IEEEabrv,IEEEexample}
\balance
\end{document}